\documentclass[conference]{IEEEtran}
\IEEEoverridecommandlockouts
\usepackage{cite}
\usepackage{amsmath,amssymb,amsfonts}
\usepackage{algorithmic}
\usepackage{graphicx}
\usepackage{textcomp}
\usepackage{xcolor}
\def\BibTeX{{\rm B\kern-.05em{\sc i\kern-.025em b}\kern-.08em
    T\kern-.1667em\lower.7ex\hbox{E}\kern-.125emX}}
\begin{document}

\title{Large Language Models for Toxic Language Detection in Low-Resource Balkan Languages\\
}

\author{\IEEEauthorblockN{1\textsuperscript{st} Amel Muminovic, \IEEEmembership{Member, IEEE}}
\IEEEauthorblockA{\textit{Faculty of Engineering} \\
\textit{International Balkan University}\\
Skopje, North Macedonia \\
amel.muminovic@ibu.edu.mk}
\and
\IEEEauthorblockN{2\textsuperscript{nd} Amela Kadric Muminovic}
\IEEEauthorblockA{\textit{School of Electrical Engineering} \\
\textit{University of Belgrade}\\
Belgrade, Serbia \\
ka243096m@student.etf.bg.ac.rs}
}

\maketitle

\begin{abstract}
Online toxic language causes real harm, especially in regions with limited moderation tools. In this study, we evaluate how large language models handle toxic comments in Serbian, Croatian, and Bosnian, languages with limited labeled data. We built and manually labeled a dataset of 4,500 YouTube and TikTok comments drawn from videos across diverse categories, including music, politics, sports, modeling, influencer content, discussions of sexism, and general topics. We demonstrate how adding minimal context can improve toxic language detection in low-resource settings and suggest practical strategies such as improved prompt design and threshold calibration. Four models (GPT-3.5 Turbo, GPT-4.1, Gemini 1.5 Pro, and Claude 3 Opus) were tested in two modes: zero-shot and context-augmented. We measured precision, recall, F1 score, accuracy and false positive rates. Including a short context snippet raised recall by about 0.12 on average and improved F1 score by up to 0.10, though it sometimes increased false positives. The best balance came from Gemini in context-augmented mode, reaching an F1 score of 0.82 and accuracy of 0.82, while zero-shot GPT-4.1 led on precision and had the lowest false alarms. These results show that prompt design alone can yield meaningful gains in toxicity detection for underserved Balkan language communities.
\end{abstract}

\begin{IEEEkeywords}
Large Language Models, Natural Language Processing, Artificial Intelligence, Toxic Language
\end{IEEEkeywords}

\section{Introduction}
Any communication that uses insults, threats, harassment or other demeaning remarks aimed at an individual or group qualifies as toxic language. This category includes crude slurs and explicit threats as well as sarcastic jabs and demeaning jokes. We adopt the definition of toxicity from Davidson et al. (2017)\cite{b1} which defines toxic language as behavior that causes emotional or psychological harm regardless of whether it targets personal traits opinions or social identities.

\subsection{Background \& Problem Statement}
Toxic language remains one of the most pressing challenges for social platforms, with significant psychological and societal impacts, particularly in politically sensitive or post-conflict regions \cite{b2}. The problem is amplified in online spaces due to anonymity, lack of accountability, and the speed at which harmful content can spread. These factors create environments where users feel emboldened to express hate or harassment they might avoid in offline settings \cite{b3}. While tech companies have invested in automated moderation systems, these systems often perform well only in high-resource languages like English \cite{b4}. Most rely on keyword filters or classifiers trained on large datasets, but they fall short in low-resource languages due to limited data, cultural nuance, and evolving online language patterns \cite{b5}. For languages spoken across the Western Balkans, especially Serbian, Bosnian, and Croatian, moderation tools are less reliable, allowing toxic language and harassment to spread with limited oversight \cite{b6}. 

A 2021 investigation by the Balkan Investigative Reporting Network (BIRN) found that nearly half of the toxic language posts reported in Balkan languages remained online, even after Facebook and Twitter confirmed that the content violated their rules \cite{b7}. Other studies have similarly highlighted systemic moderation shortcomings across smaller linguistic communities, particularly emphasizing the limited training data and cultural context necessary for effective AI moderation of toxic language \cite{b8}. These moderation challenges in the Balkans are further complicated by historical tensions, entrenched political polarization, and specific cultural sensitivities, which significantly impact how toxicity manifests and is perceived within these communities \cite{b9}.

Platforms like YouTube and TikTok play an important role in shaping online conversations in these regions \cite{b10, b11}. However, the informal tone of comment sections, frequent code-switching between Latin and Cyrillic scripts, and the use of slang or sarcasm make harmful content harder to detect automatically \cite{b12}. Large language models (LLMs) such as GPT, Claude, and Gemini offer new moderation possibilities thanks to their generalization capabilities and multilingual support. Yet, their performance in low-resource, context-heavy environments remains underexplored \cite{b13}.

This gap raises an important question about whether simple improvements, such as adding brief contextual information, could make these models more effective in practice. This paper examines how the performance of large language models changes when detecting toxic language in Balkan language comments, first using zero-shot prompts and then by adding minimal context. We test whether a short description of the video a comment responds to can help the model better interpret intent and tone. Our hypothesis is that even a small amount of context can help models distinguish between ordinary criticism and toxic remarks, especially when comments involve cultural references, sarcasm, or platform-specific slang.

To our knowledge, no prior study has systematically evaluated the performance of state-of-the-art large language models for toxic language detection in Serbian, Croatian, and Bosnian, making this a notable gap in the current research landscape.

\subsection{Objectives and Contributions}

The goal of this paper is to evaluate how the presence of contextual information affects the ability of large language models to detect toxic language in comments written in Balkan languages. We focus on three languages: Serbian, Croatian, and Bosnian, where reliable labeled datasets are limited and where linguistic variation makes moderation difficult. Specifically, we compare zero-shot prompts with and without additional context, such as a brief description of the video being commented on.

Our main contributions are:
\begin{itemize}
    \item Introducing a manually labeled dataset of 4\,500 YouTube and TikTok comments (1\,500 per language), spanning a range of categories including music, politics, sports, modeling, influencers, sexism, and general topics.
    
    \item Evaluating multiple versions of state-of-the-art language models (GPT-3.5 Turbo, GPT-4.1, Gemini 1.5 Pro and Claude 3 Opus), comparing their performance with and without added context, and analyzing how contextual prompts affect false negatives across languages.
    
    \item Providing qualitative insights into common failure cases, including misinterpretation of slang, sarcasm, and mixed script comments, which highlight the limits of zero-shot moderation in low resource settings.
\end{itemize}

\section{Related work}

\subsection{Toxic Language Detection}

Prior work on toxic language detection has focused on various machine learning approaches and challenges, especially in high-resource languages. A variety of classifiers such as Support Vector Machines (SVM), Random Forests, and deep learning architectures like Convolutional Neural Networks (CNNs) and Long Short-Term Memory networks (LSTMs) have been successfully applied to this problem \cite{b14, b15}.

Recent advancements in deep learning and transformer-based architectures have significantly improved toxic language detection. Transformer models, especially multilingual variants such as XLM-RoBERTa and multilingual BERT, have exhibited notable improvements due to their ability to capture contextual nuances more effectively than traditional methods \cite{b16}, \cite{b17}. However, despite their strong performance, transformer models still struggle with low-resource languages due to limited training data and the complexity of informal, regionally diverse, and culturally specific language use \cite{b18}.

\subsection{Large Language Model-based Methods}

In recent years, large language models, such as GPT‑3, GPT‑4, Claude, and Gemini, have increasingly been leveraged for text classification tasks, including the identification of toxic language. These models have been particularly effective in zero‑shot and few‑shot scenarios, enabling them to generalize and perform well even with limited task‑specific annotated data \cite{b19}. Studies involving GPT‑based models have demonstrated their versatility in handling various types of toxic and abusive content, showcasing strong generalization across multiple tasks and datasets, albeit primarily in English‑centric evaluations \cite{b20, b21}.

Innovations in prompting strategies, such as chain-of-thought (CoT) prompting, further enhance LLM capabilities by guiding models to reason step by step, which has shown strong improvements in reasoning tasks \cite{b22}. Recent work has demonstrated that CoT-based frameworks, such as HateGuard, can significantly improve the detection of emerging and contextually complex hate speech in zero-shot settings \cite{b23}. However, despite this progress, the broader effectiveness of CoT prompting in detecting nuanced toxic language, particularly in cases that involve cultural references, sarcasm, or code switching, remains insufficiently studied. Existing research highlights these limitations. For example, LLMs still struggle with covert toxicity, including irony and coded insults, which reveals ongoing gaps in reliably identifying subtle forms of harm \cite{b24}. We do not experiment with CoT in this paper, but this suggests an important direction for future research.

\subsection{Low-Resource Languages and Natural Language Processing}

Processing natural language effectively in low-resource languages presents numerous challenges, notably due to the scarcity of extensive annotated datasets necessary for training robust machine learning models. These challenges are further amplified by the complex nature of many low-resource languages, which often include rich grammar, regional dialects, multiple writing systems, and frequent use of slang and idiomatic expressions \cite{b25}. In the case of Balkan languages such as Serbian, Bosnian, and Croatian, the combination of morphological complexity, dialectal diversity, and script variation (e.g., Serbian’s use of both Cyrillic and Latin) introduces additional hurdles for Natural Language Processing (NLP) pipelines \cite{b26, b27}.

Recent work by Bogdanović et al. (2024) on the development of SRBerta, a transformer-based model trained on Serbian Cyrillic legal texts, highlights the potential for domain-specific LLMs in low-resource settings while also underscoring the need for substantial annotated corpora \cite{b28}. Complementing this, Drasković et al. (2022) demonstrated the challenges of sentiment classification in Serbian using multilingual datasets and traditional classifiers, further emphasizing the importance of both data quality and language-specific modeling strategies \cite{b29}.

Moreover, regional studies such as those by Milaković et al. (2024) underline ongoing efforts and limitations in developing practical NLP tools for toxic content moderation in Serbian, emphasizing both the urgent need and existing barriers for comprehensive NLP solutions in Balkan languages \cite{b30}. Although there have been promising initial explorations in the region, such as Kopani and Llapushi’s (2025) study on fine-tuning GPT-3.5 for detecting hate speech in Albanian YouTube comments, comprehensive empirical evaluation and benchmarking across Serbian, Bosnian, and Croatian remain notably sparse \cite{b31}.

This study expands on the above findings to empirically evaluate large language models on toxic comment detection in Balkan languages.

\section{Methodology}
\begin{figure*}[htbp]
  \centering
  \includegraphics[width=\textwidth]{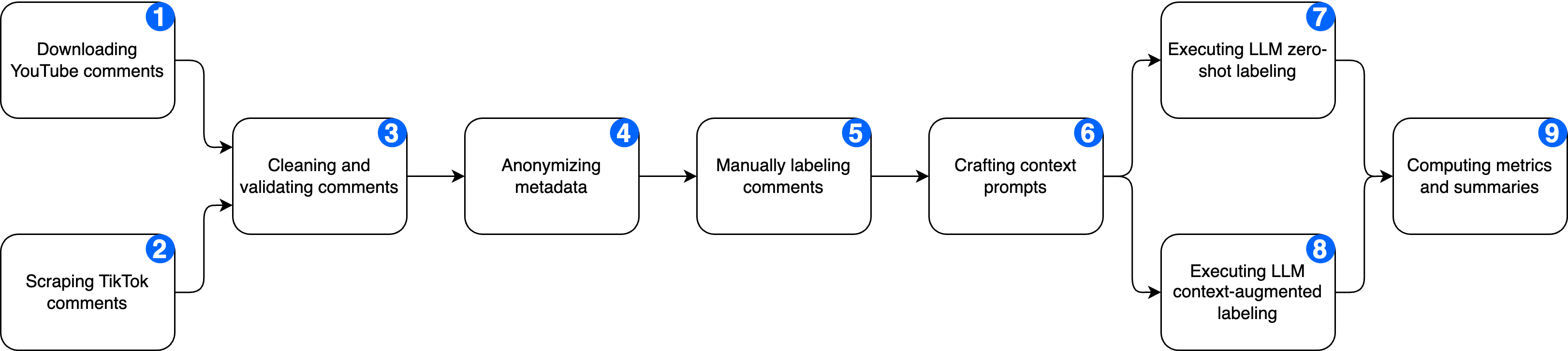}
  \caption{End-to-end pipeline for our toxicity detection study. Steps 1 and 2 show data collection from YouTube and TikTok; steps 3–6 cover cleaning, anonymization, manual labeling, and context prompt crafting; steps 7–8 execute the large language models in zero-shot and context-augmented modes; and step 9 computes all evaluation metrics.}
  \label{fig:pipeline}
\end{figure*}

The methodology for this study follows a structured multi-step pipeline designed to collect, process, and evaluate user comments for toxic language detection in low-resource Balkan languages. The process begins with data acquisition from social media platforms, followed by data cleaning and manual annotation. Contextual information is then incorporated to evaluate prompting strategies with large language models. Model outputs are subsequently analyzed using quantitative metrics. An overview of the methodology pipeline is provided in Figure~\ref{fig:pipeline}.

\subsection{Dataset Collection}
Building on the identified need for benchmark datasets in low-resource Balkan languages, we constructed a multilingual dataset of 4\,500 user comments, with 1\,500 in each of the three target languages: Serbian, Bosnian, and Croatian.

Comments were collected from publicly available videos on YouTube and TikTok via the YouTube Data API and a comment scraping script for TikTok. Approximately 66.6\% of the data comes from YouTube and 33.3\% from TikTok. These platforms were chosen due to their large user bases and frequent comment activity, which make them suitable sources for observing both explicit and subtle forms of toxicity.

In order to reflect a wide range of online discourse, videos were selected from four content categories: politics, music, sports, and general/pop culture. These categories were chosen to ensure diversity in the dataset and to capture different tones and contexts where toxic language may appear. Videos with high engagement and visible controversy were prioritized in the selection to increase the presence of strong opinions and emotionally charged comments. These data collection steps correspond to steps 1 and 2 in Figure~\ref{fig:pipeline}.

\subsection{Dataset Annotation}
The annotation of the dataset was conducted manually by the two authors of this paper. Each annotator was familiar with the video content from which the comments were sourced, allowing for context-informed judgments. Every comment was labeled as either Toxic or Non-Toxic. The labeling was guided by the definition of toxic language adopted from Davidson et al.~\cite{b1}, which considers toxic language to be any communication that causes emotional or psychological harm, including insults, threats, harassment, demeaning remarks, crude slurs, sarcastic jabs, or jokes aimed at individuals or groups.

Annotation was performed using a shared spreadsheet, which also served for data management and conflict tracking. The annotators independently labeled all comments, reaching agreement on 94\% of the cases. Disagreements were resolved through discussion until consensus was reached. Inter-annotator agreement was measured using Cohen’s Kappa coefficient, yielding a value of 0.87, indicating strong reliability. Labels were not exposed to the language models at any stage of the evaluation process. Dataset annotation aligns with step 5 depicted in Figure~\ref{fig:pipeline}.

\subsection{Data Preparation}
To ensure quality and privacy, all comments were preprocessed before model evaluation. Comments consisting solely of emojis or entirely in non-target languages were excluded to maintain linguistic relevance. Preprocessing involved UTF-8 decoding, trimming unnecessary whitespace, and preserving emojis and punctuation, as these often convey important cues in online discourse. Personally identifiable information such as user IDs, comment IDs, and video IDs was not shared with the models. Mentions of public figures were retained during inference to preserve contextual meaning but were anonymized afterward to allow ethical data sharing. 

For each of the 15 videos, a short context was manually crafted, describing the video's content and individuals featured, later used in context-augmented prompting. We deliberately opted for short, hand-crafted context snippets, rather than passing generic video metadata such as the title or description, to evaluate whether focused, concrete context could meaningfully assist model disambiguation. While this manual approach is not scalable for large-scale moderation, our intent was to isolate the effect of targeted context. In practice, scalable alternatives exist: automated context extraction tools, such as LLM-driven summarization or topic modeling, could generate similar prompts in production systems. Future work should explore the trade-offs between these automated methods and human-authored context. 

This preprocessing and context crafting correspond to steps 3, 4, and 6 in Figure~\ref{fig:pipeline}.

\subsection{Experimental Design}
To evaluate the performance of large language models on toxic language detection in Serbian, Bosnian, and Croatian, we tested four state-of-the-art models: OpenAI GPT-3.5 Turbo, OpenAI GPT-4.1, Google Gemini 1.5 Pro, and Anthropic Claude 3 Opus. These models were selected based on their accessibility through official APIs, strong performance on multilingual and safety-related benchmarks, and their widespread usage in real-world NLP applications. All inference was performed with a temperature setting of 0 and a maximum token limit of 10 to ensure deterministic and concise outputs.

We compared model behavior across two prompting strategies: zero-shot (where the model receives only the comment and a task description) and context-augmented (where the model also receives a short description of the video to help interpret the comment). In the zero-shot scenario, the prompt presented the model with a user comment to determine toxicity:

\begin{quote}\small
\texttt{Analyze the following comment written in Serbian, Bosnian, or Croatian. The comment is a response to a YouTube or TikTok video. Determine if the comment contains toxic language, including insults, hate speech, threats, harassment, or harmful sarcasm. Consider regional slang, sarcasm, tone, and cultural context. Respond with 0 if the comment is not toxic, and 1 if it is toxic. Do not respond with any other text. Comment is: <COMMENT\_TEXT>}
\end{quote}

In the context-augmented setup, models were additionally provided with a manually crafted context about the video's content and featured individuals:

\begin{quote}\small
\texttt{Analyze the following comment written in Serbian, Bosnian, or Croatian. The comment is a response to a YouTube or TikTok video. Use the provided context to understand what the video is about. Context about video: <CONTEXT> Determine if the comment contains toxic language, including insults, hate speech, threats, harassment, or harmful sarcasm. Consider regional slang, sarcasm, tone, and cultural context. Respond with 0 if the comment is not toxic, and 1 if it is toxic. Do not respond with any other text. Comment is: <COMMENT\_TEXT>}
\end{quote}

All model outputs adhered strictly to the binary classification format (0 for non-toxic, 1 for toxic). This evaluation stage corresponds to steps 7 and 8 in Figure~\ref{fig:pipeline}.

\subsection{Evaluation Metrics}
To assess model performance in classifying toxic versus non-toxic comments, we used standard classification metrics: precision, recall, F1 score, accuracy and false positive rate (FPR). These metrics are commonly applied in NLP classification tasks to capture different aspects of prediction quality. Our dataset contained approximately 51\% toxic and 49\% non-toxic comments, allowing for a reliable application of this metric without requiring additional adjustments for class imbalance. We computed all metrics across the full dataset, treating Serbian, Bosnian, and Croatian jointly due to their high linguistic similarity. In addition, we reported F1 scores separately for each language to highlight any performance differences between them. The computation of these metrics is step 9 in Figure~\ref{fig:pipeline}. In the following section, we present and compare the results of all models under both zero-shot and context-augmented setups.

\section{Results}

\subsection{Overview of Model Performance}

Table~\ref{tab:results} summarizes model performance on toxicity detection in low-resource languages, in both zero-shot and context-augmented configurations. We assign each setup a short label (for example, GPT-3.5 Turbo zero-shot (G3-Z)) to keep figures and the text uncluttered. Overall, accuracy ranges from 0.734 for Claude 3 Opus zero-shot (CL-Z) up to 0.823 for Gemini 1.5 Pro context-augmented (GM-C), suggesting that context-augmentation often yields a measurable boost.

Beyond raw accuracy, the counts of true positives, false positives, true negatives, and false negatives reveal how each system balances the tension between over-flagging and under-flagging toxic comments. For instance, GPT-4.1 zero-shot (G4-Z) flags only 86 benign comments as toxic but misses 962 actual toxic instances. By contrast, GPT-3.5 Turbo context-augmented (G3-C) catches more toxic content, reducing its false negatives from 811 to 480, though its false positives climb from 332 to 467.

\begin{table*}[htbp]
  \centering
  \caption{Model Performance on Toxicity Detection in Low-Resource Languages by Configuration}
  \label{tab:results}
  \begin{tabular}{|c|l|c|c|c|c||c|c|c|c|c|}
    \hline
    \textbf{Label} & \textbf{Model}
      & \multicolumn{4}{c||}{\textbf{Counts}}
      & \multicolumn{5}{c|}{\textbf{Metrics}} \\
    \cline{3-11}
    & & TP & TN & FP & FN
      & Accuracy & Precision & Recall
      & F1 score & FPR \\
    \hline
    G3-Z & gpt-3.5 turbo zero-shot
      & 1\,489 & 1\,868 & 332  & 811
      & 0.746 & 0.818 & 0.647 & 0.723 & 0.151 \\ \hline
    G3-C & gpt-3.5 turbo context-augmented
      & 1\,820 & 1\,733 & 467  & 480
      & 0.790 & 0.796 & \textbf{0.791} & 0.794 & 0.212 \\ \hline
    G4-Z & gpt-4.1 zero-shot
      & 1\,338 & 2\,114 &  86  & 962
      & 0.767 & \textbf{0.940} & 0.582 & 0.719 & \textbf{0.039} \\ \hline
    G4-C & gpt-4.1 context-augmented
      & 1\,592 & 2\,095 & 105  & 708
      & 0.819 & 0.938 & 0.692 & 0.797 & 0.048 \\ \hline
    GM-Z & gemini-1.5-pro zero-shot
      & 1\,388 & 2\,001 & 199  & 912
      & 0.753 & 0.875 & 0.603 & 0.714 & 0.090 \\ \hline
    GM-C & gemini-1.5-pro context-augmented
      & 1\,806 & 1\,897 & 303  & 494
      & \textbf{0.823} & 0.856 & 0.785 & \textbf{0.819} & 0.138 \\ \hline
    CL-Z & claude-3-opus zero-shot
      & 1\,188 & 2\,117 &  83  & 1\,112
      & 0.734 & 0.935 & 0.517 & 0.665 & 0.038 \\ \hline
    CL-C & claude-3-opus context-augmented
      & 1\,440 & 2\,035 & 165  & 860
      & 0.772 & 0.897 & 0.626 & 0.738 & 0.075 \\ \hline
      \multicolumn{11}{|l|}{$^{\mathrm{a}}$Bold values indicate best results.} \\
\hline

  \end{tabular}
  
\end{table*}

\subsection{Precision versus Recall}

\begin{figure}[htbp]
  \centerline{\includegraphics[width=0.48\textwidth]{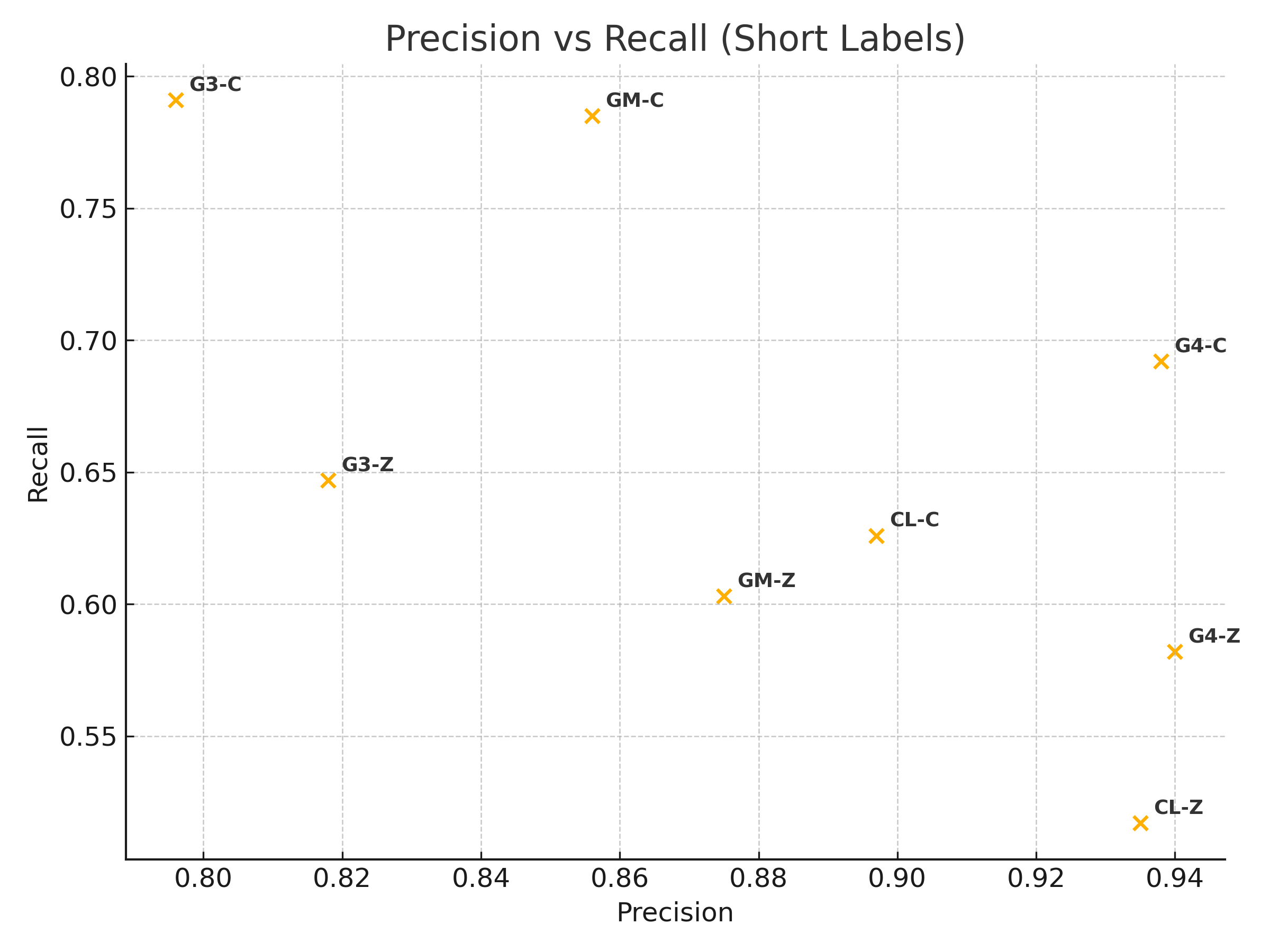}}
  \caption{Precision versus recall for each model configuration; full model names and labels defined in each paragraph.}
  \label{fig:precision_vs_recall}
\end{figure}

Figure~\ref{fig:precision_vs_recall} plots precision against recall for every configuration. GPT-4.1 zero-shot (G4-Z) achieved a precision of 0.940, meaning that fewer than six percent of its flagged comments were false alarms, but its recall of 0.582 means it missed over forty percent of toxic remarks. In many real-world settings where missing harmful content is more detrimental than a few extra false positives, a higher recall is preferable.

Context-augmented variants such as GPT-3.5 Turbo context-augmented (G3-C) and Gemini 1.5 Pro context-augmented (GM-C) raise recall into the high seventies while keeping precision above 0.79. That balance makes them more suitable for applications demanding both sensitivity and reliability.

\subsection{False Positive Rate versus Recall}

\begin{figure}[htbp]
  \centerline{\includegraphics[width=0.48\textwidth]{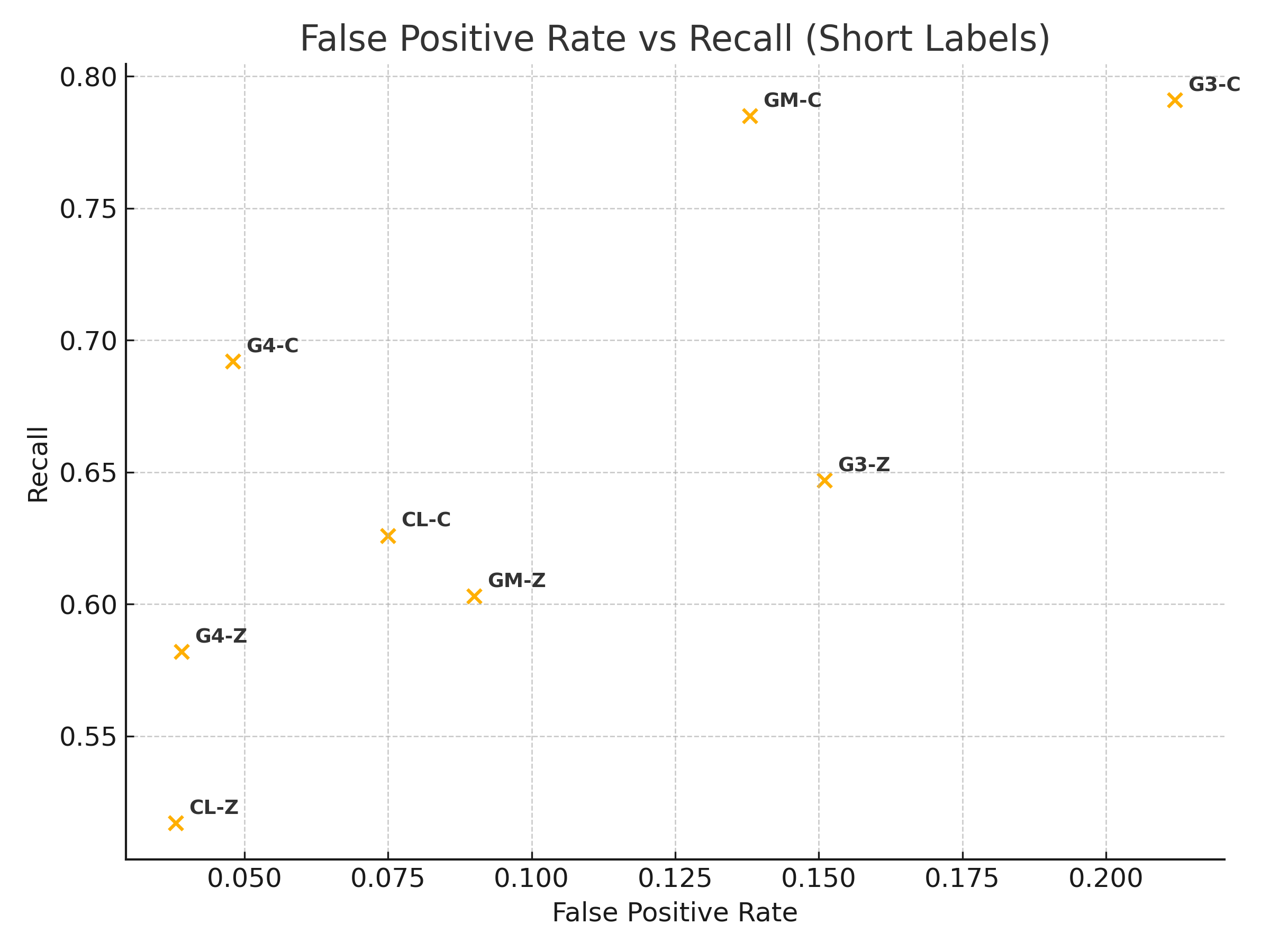}}
  \caption{False positive rate versus recall for each configuration; full model names and labels defined in each paragraph.}
  \label{fig:fpr_vs_recall}
\end{figure}

To further illustrate the trade-off, Figure~\ref{fig:fpr_vs_recall} shows false positive rate against recall. GPT-4.1 zero-shot (G4-Z) and Claude 3 Opus zero-shot (CL-Z) appear closest to the ideal top-left region, demonstrating they rarely mislabel benign comments while still catching a fair share of toxicity. By contrast, GPT-3.5 Turbo context-augmented (G3-C) reaches a recall of 0.791 but at the cost of an FPR of 0.212, indicating more than one in five benign comments gets flagged.

\subsection{Importance of Context}

\begin{figure}[htbp]
  \centerline{\includegraphics[width=0.48\textwidth]{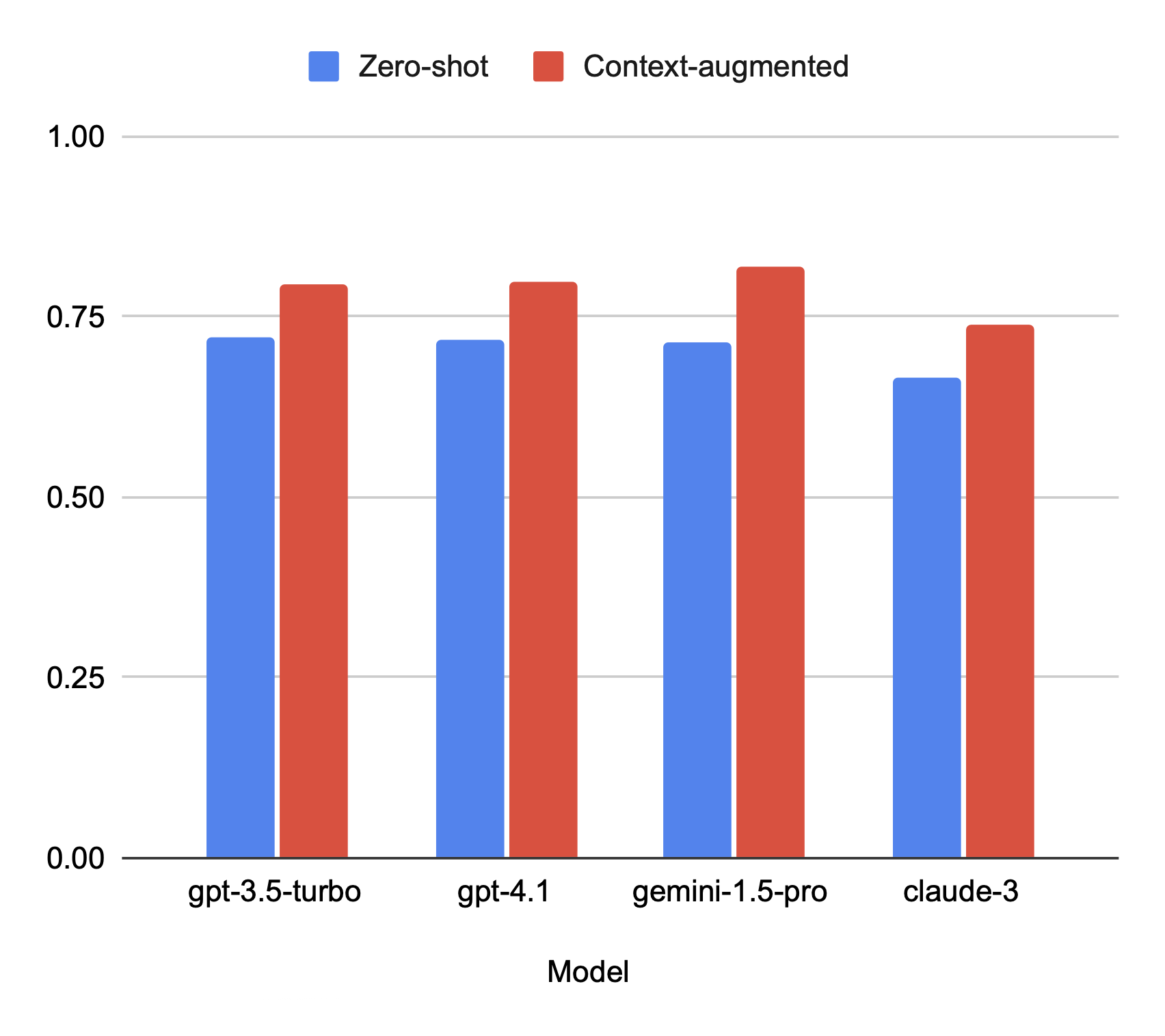}}
  \caption{F1 scores for zero-shot and context-augmented configurations across models}
  \label{fig:f1_scores}
\end{figure}

Figure~\ref{fig:f1_scores} presents a bar chart of F1 scores for each model in zero-shot and context-augmented modes. GPT-3.5 Turbo zero-shot (G3-Z) has an F1 score of 0.723, which increases to 0.794 for GPT-3.5 Turbo context-augmented (G3-C). Similarly, GPT-4.1 zero-shot (G4-Z) rises from 0.719 to 0.797 when context is added, while Gemini 1.5 Pro zero-shot (GM-Z) goes from 0.714 to 0.819 for Gemini 1.5 Pro context-augmented (GM-C), and Claude 3 Opus zero-shot (CL-Z) from 0.665 to 0.738 for Claude 3 Opus context-augmented (CL-C). Notably, GM-C achieves the highest F1 score of 0.819, reflecting its strong balance between precision and recall.

\subsection{Cost Comparison}
We measured API costs for 500 requests and then scaled to all 4\,500 comments in our dataset (9 batches of 500), under both zero-shot and context-augmented modes. Figures assume list prices at the time of writing and a temperature setting of 0. Table~\ref{tab:cost-comparison} shows total API costs for 500 and 4\,500 requests under each configuration.

\begin{table}[htbp]
  \centering
  \caption{API cost comparison for 500 and 4\,500 requests}
  \label{tab:cost-comparison}
  \footnotesize
  \begin{tabular}{|l|c|c|c|c|}
    \hline
    \textbf{Model} 
      & \multicolumn{2}{c|}{\textbf{Cost for 500}} 
      & \multicolumn{2}{c|}{\textbf{Cost for 4\,500}} \\
    \cline{2-5}
      & Zero-shot 
      & \shortstack{Context-\\augmented} 
      & Zero-shot 
      & \shortstack{Context-\\augmented} \\
    \hline
    GPT-3.5 Turbo  & \$0.0391  & \$0.0515  & \$0.3520  & \$0.4634 \\
    \hline
    GPT-4.1        & \$0.1481  & \$0.1971  & \$1.3329  & \$1.7739 \\
    \hline
    Gemini 1.5 Pro & \$0.0929  & \$0.1241  & \$0.8363  & \$1.1168 \\
    \hline
    Claude 3 Opus  & \$1.4441  & \$1.8341  & \$12.9969 & \$16.5069 \\
    \hline
  \end{tabular}
\end{table}

GPT-3.5 Turbo is the most cost-effective model, balancing cost and performance well. Claude 3 Opus, however, is significantly more expensive without proportional performance improvements. Context augmentation increases costs by about 30\%–40\% per model. Platforms should thus carefully weigh incremental performance gains against increased costs, using context augmentation selectively for ambiguous or critical moderation tasks.

\subsection{Summary of Findings}

Context augmentation consistently boosted recall and F1 scores by roughly 0.07 to 0.10, although it sometimes led to higher false positive rates. The most balanced performer is Gemini 1.5 Pro context-augmented (GM-C), which achieved both the highest F1 score of 0.819 and the top accuracy of 0.823. GPT-4.1 zero-shot (G4-Z) remains the precision leader, making it suitable when false alarms carry a high penalty.

These results underscore that the optimal model configuration depends on the specific requirements of a moderation task, whether that’s minimizing false positives, maximizing detection of toxicity, or finding a middle ground. In the next section, we discuss how calibration, threshold tuning, and simple ensemble approach can further refine these trade-offs.

\subsection{Performance by Language}

We evaluated F1 scores for Serbian, Bosnian, and Croatian separately to see if model performance varied by language. Table~\ref{tab:lang-f1} shows results for each model and configuration. In every case, context-augmentation improved F1 scores across all three languages. Gemini 1.5 Pro context-augmented achieved the highest scores overall, with particularly strong gains for Croatian. Performance was consistent, showing that the context-augmented approach generalizes well across these closely related languages.

\begin{table}[htbp]
  \centering
  \caption{F1 Score by Language and Model}
  \label{tab:lang-f1}
  \footnotesize
  \begin{tabular}{|l|c|c|c|}
    \hline
    \textbf{Model} & \textbf{Serbian } & \textbf{Bosnian} & \textbf{Croatian} \\
    \hline
    GPT-3.5 Turbo zero-shot              & 0.725 & 0.721 & 0.722 \\
    \hline
    GPT-3.5 Turbo context-augmented      & 0.791 & 0.784 & 0.806 \\
    \hline
    GPT-4.1 zero-shot              & 0.747 & 0.732 & 0.673 \\
    \hline
    GPT-4.1 context-augmented      & 0.811 & 0.782 & 0.798 \\
    \hline
    Gemini 1.5 Pro zero-shot           & 0.729 & 0.709 & 0.704 \\
    \hline
    Gemini 1.5 Pro context-augmented   & 0.822 & 0.806 & 0.830 \\
    \hline
    Claude 3 Opus zero-shot             & 0.675 & 0.679 & 0.640 \\
    \hline
    Claude 3 Opus context-augmented     & 0.752 & 0.746 & 0.713 \\
    \hline
  \end{tabular}
\end{table}

Overall, these results show that context-augmented prompting benefits all three languages, with Gemini 1.5 Pro context-augmented giving the strongest performance.

\section{Discussion}

Our results show that even a small amount of context can substantially improve the sensitivity of large language models in detecting toxic content in low-resource Balkan languages. Below, we highlight the key takeaways, practical implications, limitations of our study, and directions for future work.

\subsection{Key Findings}

First, context-augmentation consistently increased recall by an average of twelve percentage points across all models, confirming our hypothesis that brief descriptions of the video or post help disambiguate cultural references, slang, and sarcasm. For instance, the comment:

\begin{quote}
“Zašto su birali bolesnu osobu?”\\
(“Why did they elect a sick person?”)
\end{quote}

was labeled toxic by our human annotators but missed by every model in zero-shot mode. After adding the context that this comment referred to a Serbian politician struggling during an interview, all context-augmented runs correctly flagged it as toxic.

Moreover, across our 4\,500-comment test set, GPT-3.5 Turbo context-augmentation reduced the number of missed toxic comments by 331 (false negatives dropped from 811 to 480). GPT-4.1 saw 254 fewer misses (962 to 708), Gemini 1.5 Pro achieved 418 fewer (912 to 494), and Claude 3 Opus had 252 fewer (1\,112 to 860).  

\subsection{Practical Implications}

These findings suggest several straightforward steps for moderation systems:

\begin{itemize}
  \item \textbf{Prompt design matters.}  Even a simple two-sentence context prompt can turn hundreds of “invisible” toxic comments into detectable ones.
  \item \textbf{Edge-case awareness.}  Context can sometimes over-flag benign content.  For example, the harmless remark:
  \begin{quote}
    “Trebao je dobiti status branitelja za ovaj pothvat.”\\
    (“He should have received veteran status for this undertaking.”)
  \end{quote}
 was made in response to a football match video, referring to a foul that stopped a counter-attack which could have led to a goal and secured a win for Serbia. Yet, GPT-3.5 Turbo context-augmented and Gemini 1.5 Pro context-augmented both misclassified it as toxic after seeing that background.
  \item \textbf{Threshold calibration and ensemble approach.}  Combining a high-precision zero-shot model (e.g., GPT-4.1 zero-shot) with a high-recall context-augmented model (e.g., GPT-3.5 Turbo context-augmented) can recover even more toxic content without unduly raising false positives.
\end{itemize}

\subsection{Limitations}

While encouraging, our study has several limitations to keep in mind:

\begin{itemize}
  \item \textbf{Dataset size and domain.} We labeled 4\,500 comments drawn from YouTube and TikTok posts across a range of categories, including music, politics, sports, modeling, influencers, sexism, and general topics. While this dataset covers diverse public discussions, results may not generalize to private messaging platforms or comments in less-represented categories.
  \item \textbf{Static context.} We provided only a short, manually written description of the video. For example, in one video featuring a well-known TikToker who was throwing money on the street, several comments unexpectedly focused on his hairstyle. This detail was not included in our context prompt, as it seemed irrelevant at the time. As a result, the models missed or misclassified some of these comments. Automated context extraction, such as summarizing user profiles, conversation threads, or external news links might help capture those unanticipated topics and further improve performance.
  \item \textbf{Limited annotation diversity.} Manual labeling was conducted by the two authors, both highly familiar with Balkan languages and culture. Though yielding strong agreement, this familiarity might introduce subtle biases. Broader annotation involving diverse annotators might better identify nuanced or culturally ambiguous cases.

  \item \textbf{Language coverage.}  We focused on Serbian, Croatian, and Bosnian. Other Balkan languages and dialects (for example, Macedonian or Albanian) may require different handling, especially when scripts or grammar differ more significantly.
\end{itemize}

\subsection{Future Work}

\begin{itemize}
  \item \textbf{Dynamic context generation.}  Investigate whether automatically generated summaries or translations into a high-resource pivot language (for example, English) can match or exceed the gains of manually written context.
  \item \textbf{Few-shot and chain-of-thought prompting.}  Experiment with few-shot examples or chain-of-thought prompts to see if guiding the model’s reasoning can reduce covert toxicity errors, such as sarcasm or coded insults.
  \item \textbf{Fine-tuning lightweight adapters.}  Explore domain-specific fine-tuning of small adapter layers on top of base LLMs, using our 4\,500-comment dataset. That may yield further improvements with minimal compute.
  \item \textbf{User-in-the-loop systems.}  Design human-review interfaces that leverage model confidence scores and context to surface borderline cases to moderators, thereby combining automated speed with human judgment.
\end{itemize}

In summary, our discussion underscores that context augmentation is a low-barrier, high-impact technique for improving toxicity detection in under-served languages. With further tuning and system integration, platforms can deploy these simple prompting strategies to make online spaces safer and more inclusive for Balkan language communities.

\section{Conclusion}

In this work, we systematically evaluated the impact of brief context prompts on the ability of large language models to detect toxic language in low-resource Balkan languages. Our experiments on a manually labeled dataset of 4\,500 YouTube and TikTok comments in Serbian, Croatian, and Bosnian showed that:

\begin{itemize}
  \item \textbf{Context-augmentation reliably boosts recall and F1 score.} Across all models, adding a two-sentence description of the video increased recall by 0.12 on average and improved F1 scores by up to 0.10, with Gemini 1.5 Pro context-augmented achieving the top F1 score of 0.819.
  \item \textbf{Trade-offs remain.} Zero-shot GPT-4.1 achieved the highest precision (0.940) and lowest false positive rate (0.039), making it preferable for settings intolerant of false alarms. Context-augmented variants closed the gap on precision while dramatically reducing missed toxic instances.
  \item \textbf{Real-world scale.} For GPT-3.5 Turbo alone, context-augmentation cut false negatives by 331 comments, hundreds of toxic remarks that would otherwise go unnoticed.
  \item \textbf{Edge cases highlight limits.} Manually written context sometimes missed unexpected comment themes (e.g. rebuttals of a sports foul), leading to over-flagging. Automating or broadening context extraction could address these gaps.
\end{itemize}

These findings demonstrate that simple prompt engineering, without any fine-tuning, can deliver state-of-the-art gains in toxicity detection for under-served languages. Looking forward, integrating dynamic context generation, exploring few-shot and chain-of-thought prompting, and incorporating lightweight fine-tuning or human-in-the-loop mechanisms promise further improvements. We hope this work encourages practitioners and researchers to adopt context-aware moderation pipelines and to extend these techniques to other low-resource and culturally rich language settings.  

\section*{Conflict of Interest}
The authors declare no conflict of interest.

\section*{Data Availability}
Instructions for obtaining the dataset, along with guidelines for reproducing the main experiments, are provided in the repository’s README at: https://github.com/Ammce/llm-balkan-toxicity

\end{document}